%% file: main.tex
\documentclass[10pt,twocolumn,letterpaper]{article}

\usepackage[table,xcdraw]{xcolor}
\usepackage{amssymb}
\usepackage{arydshln}
\usepackage{pifont}
\usepackage{graphicx} 
\usepackage[pagenumbers]{iccv} 


\definecolor{iccvblue}{rgb}{0.21,0.49,0.74}
\usepackage[pagebackref,breaklinks,colorlinks,allcolors=iccvblue]{hyperref}

\def\paperID{7200} 
\def\confName{ICCV}
\def\confYear{2025}

\title{Can Large Language Models Unveil the Mysteries?\\
An Exploration of Their Ability to Unlock Information in Complex Scenarios}

\author{
Chao Wang\textsuperscript{1} \hspace{1cm}
Luning Zhang\textsuperscript{1} \hspace{1cm}
Zheng Wang\textsuperscript{2} \hspace{1cm}
Yang Zhou\textsuperscript{1} \\
\textsuperscript{1}Shanghai University, Shanghai, China \hspace{1cm}
\textsuperscript{2}Zhejiang University of Technology, Hangzhou, China \\
{\tt\small \{cwang, saber\_mio, zhangluning\}@shu.edu.cn} \hspace{1cm} 
{\tt\small zhengwang@zjut.edu.cn}
}

\begin{document}
\maketitle
\input{sec/abstract}

\input{sec/introduction}
\input{sec/problem_formulation}
\input{sec/benchmark}
\input{sec/method}
\input{sec/experiments}
\input{sec/conclusion}

{
    \small
    \bibliographystyle{ieeenat_fullname}
    \bibliography{main}
}
\input{sec/appendix}

\end{document}

%% file: sec/abstract.tex
\begin{abstract}\label{sec:absract}
Combining multiple perceptual inputs and performing combinatorial reasoning in complex scenarios is a sophisticated cognitive function in humans. 
With advancements in multi-modal large language models, recent benchmarks tend to evaluate visual understanding across multiple images. 
However, they often overlook the necessity of combinatorial reasoning across multiple perceptual information. 
To explore the ability of advanced models to integrate multiple perceptual inputs for combinatorial reasoning in complex scenarios, we introduce two benchmarks: \textbf{C}lue-\textbf{V}isual \textbf{Q}uestion \textbf{A}nswering~(\textbf{CVQA}), with three task types to assess visual comprehension and synthesis, and \textbf{C}lue of \textbf{P}assword-\textbf{V}isual \textbf{Q}uestion \textbf{A}nswering~(\textbf{CPVQA}), with two task types focused on accurate interpretation and application of visual data. 
For our benchmarks, we present three plug-and-play approaches: utilizing model input for reasoning, enhancing reasoning through minimum margin decoding with randomness generation, and retrieving semantically relevant visual information for effective data integration. 
The combined results reveal current models’ poor performance on combinatorial reasoning benchmarks, even the state-of-the-art (SOTA) closed-source model achieves only 33.04\% accuracy on \textbf{CVQA}, and drops to 7.38\% on \textbf{CPVQA}. Notably, our approach improves the performance of models on combinatorial reasoning, with a 22.17\% boost on \textbf{CVQA} and 9.40\% on \textbf{CPVQA} over the SOTA closed-source model, demonstrating its effectiveness in enhancing combinatorial reasoning with multiple perceptual inputs in complex scenarios. The code will be publicly available.
\end{abstract}

%% file: sec/introduction.tex
\section{Introduction}
\label{sec:intro}
\begin{figure}
    \centering
    \includegraphics[width=\linewidth]{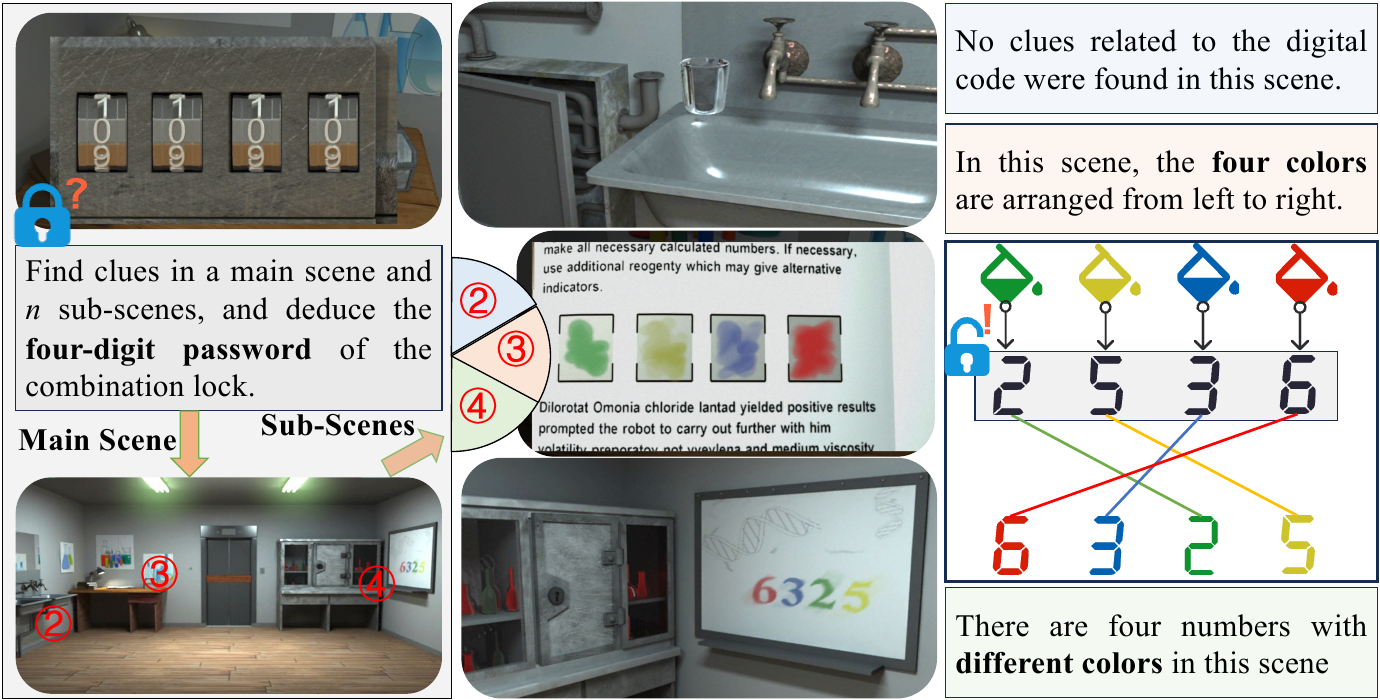}
    \caption{\textbf{Task Example.} Illustrates combinatorial reasoning with multiple perceptual inputs in complex scenarios.}
    \label{fig:intro_example}
\end{figure}

In complex scenarios, humans can integrate multiple perceptual information to perform combinatorial reasoning. As shown in Figure~\ref{fig:intro_example}, color sequences are utilized to reorder colored numbers within complex scenarios. 
In recent years, Multi-modal Large Language Models (MLLMs) \cite{chatgpt,LLaVA,Qwen-vl,Qwenmax,Geminipro} have advanced the development of visual language tasks by integrating visual encoders into pre-trained LLMs \cite{llama,Vicuna,chatgpt} to enable visual processing capabilities. 
Increasing research attention has been directed toward visual understanding in complex scenarios, including identifying positional relationships among entities~\cite{gsrbench,srbench} and reasoning in complex scenarios~\cite{Yu2020ReClorAR,Sinha2018CompositionalLU}, such as visual commonsense reasoning~\cite{vcr,GVQA,Wang2020VisualCR}, which often relies on single facts rather than integrating multiple information sources. However, prior studies have not examined models’ abilities to combine multiple perceptual inputs and perform combinatorial reasoning in complex scenarios. 

Unfortunately, there are currently no datasets or simulation scenarios that combine multiple perceptual information for combinatorial reasoning. However, we find that human often pay attention to different clues and then infer the answer when playing a game called escape room. We try to utilize ChatGPT-4o~\cite{chatgpt}, a powerful closed-source model that can perform reasoning, to conduct exploratory experiments in the game \textit{Can you escape? }~\cite{game}, as shown in Sec.~\ref{subsec:explori}, and the results failed in reasoning in 36 complex scenarios. To address the lack of evaluation in integrating multiple perceptual inputs and performing combinatorial reasoning in complex scenarios—where the intricacy and accuracy of the task make errors significantly impact the reasoning process—we meticulously curate two benchmarks with enhanced rigor. Specifically, we manually select 155 complex scenarios from this game, encompassing environments commonly encountered in real life (such as schools, laboratories, and stadiums). Furthermore, we systematically modify the pigment-related information in the images, thereby expanding the benchmark size to 2.4 times its original scale.

Specifically, we propose two benchmarks: 1) \textbf{C}lue-\textbf{V}isual \textbf{Q}uestion \textbf{A}nswering~(\textbf{CVQA}) comprises three task types designed to test a model’s ability to integrate textual and visual perceptual information to identify relevant clues from unspecified visual data. 
The first task involves reasoning with a combination of unspecified visual entities located in different scenes that can interact, such as a ``key'' in one scene that can be used to unlock a ``lock'' in another scene. 
The second task involves reasoning with specified textual entities and unspecified visual entities, where the entities mentioned in the text can interact with a certain combination of visual entities in complex scenarios. 
The third task utilizes specified sequence types (e.g., number sequences, letter sequences, etc.) to combine all visual scenes to infer clues, such as `Four colored numbers' and `Four color sequences' in Figure~\ref{fig:intro_example}. 
2) \textbf{C}lue of \textbf{P}assword-\textbf{V}isual \textbf{Q}uestion \textbf{A}nswering~(\textbf{CPVQA}) consists of two task types that evaluate the model’s ability to combine visual perceptual information for accurate combinatorial reasoning. These tasks involve reasoning across all visual scenes to determine the final numeric sequence (e.g., 2536 in Figure~\ref{fig:intro_example}) or entity code sequence (e.g., BCDA in Figure~\ref{fig:problem_formulation}).
In particular, our benchmarks include several unique features: 
1) The questions in our benchmarks are designed to guide the model to achieve combinatorial reasoning in complex scenarios. 
2) The topics in each of our complex scenarios remain consistent by selecting visually similar images from the same game level~\cite{game} rather than from existing datasets.
3) Our benchmarks' scale is manually expanded to produce high-quality evaluations, with the effectiveness of this approach demonstrated in Sec.~\ref{subsec:extension method}.

To enhance the model’s capability in integrating multiple perceptual inputs for combinatorial reasoning in complex scenes, we propose three methodological approaches from distinct perspectives: 1) From the context, contextual learning ~\cite{fewshotlearning} has been shown to be effective. We utilize text as a medium for reasoning enhancement in Sec.~\ref{subsec:LLMs and MLLMs}. 2) From the model, the probability difference in prediction logic is used to enhance reasoning ~\cite{COTwithoutprompting}, as shown in Sec.~\ref{subsec:COT without prompting}. 3) From the data, based on recent progress ~\cite{samcamsemantic,samsemantic}, we propose an image-based semantic retrieval method in Sec.~\ref{subsec:semantic and visual retrieval} to enhance the visual details of MLLMs.

We make a thorough evaluation of various well-known MLLMs and LLMs on our proposed benchmarks. In \textbf{CVQA}, results of contextual learning~(Sec.~\ref{subsec:LLMs and MLLMs}) are suboptimal, notably, Chain-Of-Thoughts~(COT) reasoning without prompting~(Sec.~\ref{subsec:COT without prompting}) and semantic retrieval~(Sec.~\ref{subsec:semantic and visual retrieval}) improve performance by up to 19\%. In \textbf{CPVQA}, performance falls short of expectations. Overall, while our method enhances combinatorial reasoning in complex scenarios, model accuracy remains low. The main problems are the tendency to answer questions based on a single source of information, over-reliance on information in the question, or reasoning based solely on the model's prior knowledge.

Our contributions can be summarized as follows: 
1) To our knowledge, this is the first study for combinatorial reasoning in complex scenarios requiring multiple perceptual inputs.
2) Experimental results indicate that even SOTA closed-source models~(such as ChatGPT-4o) fail to complete our proposed benchmark, we analysis of their errors and potential improvements. 
3) The proposed plug-and-play method achieves comparable improvements over SOTA closed-source models.

%% file: sec/problem_formulation.tex
\section{Problem Formulation}
\label{sec:formatting}
In this section, we introduce the problem formulation for the two benchmarks \textbf{CVQA} and \textbf{CPVQA}.
\subsection{Formulation of CVQA}
\label{subsec:CVQA}
\begin{figure}[ht]
    \centering
    \includegraphics[width=\linewidth]{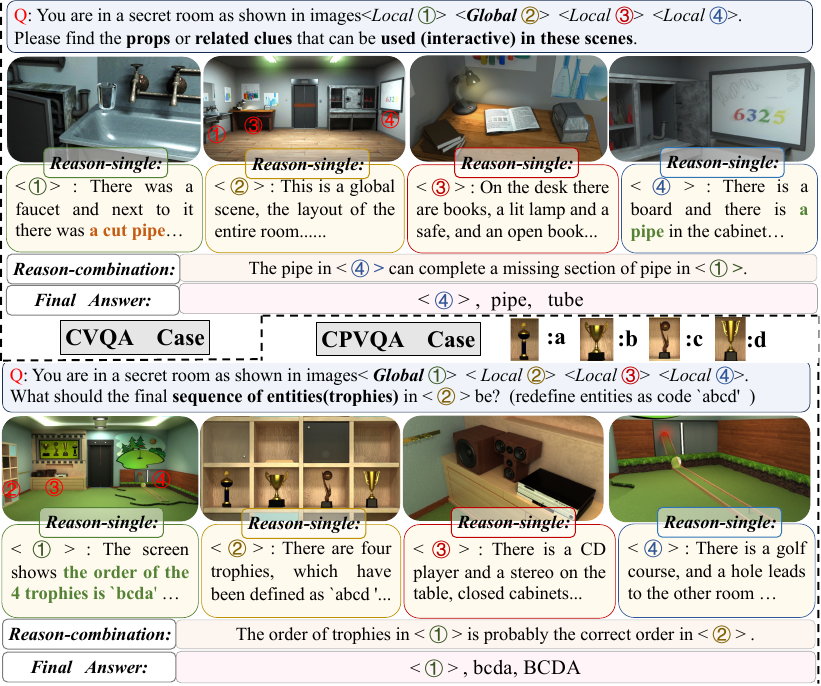}
    \caption{\textbf{Solution Formats.} The solution formats for different benchmarks in \textbf{CVQA} and \textbf{CPVQA}.}
    \label{fig:problem_formulation}
\end{figure}

Given a question text $Q_c$ and an image set \(I_c = \{i_1,i_2,\dots,i_n\}\), \textbf{CVQA} can be categorized into three types: props search, props usage, and password clues.
\begin{itemize}
    \item \textbf{Props search.} The question text $Q_{c}$ serves as a guide, directing the model to combine information in complex scenarios $I_{c}$ and identify objects that can be utilized. As illustrated in Figure~\ref{fig:problem_formulation}, the $Q_{c}$ is `\textit{Please find the props or related clues that can be used (interactive) in these scenes.}', guiding the model to infer that the half-pipe in scene \textit{4} can be applied to the missing pipe in scene \textit{1}.
    \item \textbf{Props usage.} The question text $Q_{c}$ explicitly specifies the name of the prop $t_{c}$, and infers where $t_{c}$ can be applied from complex scenarios $I_{c}$. As shown in Figure~\ref{fig:figappendix1}, the $Q_{c}$ is `\textit{And you have a prop: USB, what can this prop interact with?}', guiding the model to infer the appropriate application of $t_{c}$ based on $Q_{c}$ and the $t_{c}$ combination, specifically the computer.
    \item \textbf{Password clues.} The clues for the password, such as numeric codes or letter sequences, are explicitly provided in the question text $Q_{c}$, while clues to decode the password in complex scenarios $I_{c}$ are inferred, or the final sequence is directly given. As shown in Figure~\ref{fig:figappendix2}, $Q_{c}$ is `\textit{Reason within this scenario and derive clues about the numeric code.}', where `\textit{numeric code}' represents password clues. The model is instructed to either infer password-related clues by integrating all scenarios in $I_{c}$ (i.e., computer) or directly provide the password, in this case, `8462'.
\end{itemize}

\subsection{Formulation of CPVQA}
\label{subsec:CPVQA}
Given a question text $Q_p$ and an image set \(I_p = \{i_a,i_b,\dots,i_k\}\), \textbf{CPVQA} is divided into two categories: passwords and sequence rearrangement.
\begin{itemize}
    \item \textbf{Passwords.} The clues for the password, such as a numeric code or a sequence of letters, are explicitly provided in the problem text $Q_{p}$, while the sequence is inferred within complex scenarios $I_{p}$. As illustrated in Figure~\ref{fig:figappendix3}, $Q_{p}$ is `\textit{Reasoning in this scenario and deriving a numerical code.}', where `\textit{numeric code}' serves as the password clue. The model is then directed to infer the numeric password sequence (e.g., 8462) by combining the password sequence found on the computer with the clues in the $Q_{p}$.
    \item \textbf{Sequence rearrangement.} The problem text $Q_{p}$ utilizes codes to represent entities that need to be rearranged within a scene $i_{k}$ in $I_{p}$ , requiring the model to integrate information from all complex scenarios in $I_{p}$ to infer the final sequence code. As illustrated in Figure~\ref{fig:problem_formulation}, $Q_{p}$ encodes the trophies in $i_{k}$, directing the model to combine the trophy sequence in scene x with the encoding defined in $Q_{p}$ to infer the final sequence,  (e.g., BCDA).
\end{itemize}

%% file: sec/benchmark.tex
\section{Benchmark}
\label{sec:benchmark}
In this section, two benchmarks, \textbf{CVQA} and \textbf{CPVQA}, are introduced in detail. First, we present the structural composition of the two benchmarks in Sec.~\ref{subsec:benchmark structure}, followed by an extension of our construction method and the effectiveness of this method is demonstrated in Sec.~\ref{subsec:extension method}.
\subsection{Benchmark Structure}
\label{subsec:benchmark structure}
In this subsection, we introduce the image sources and question types of our proposed benchmarks.\\
\textbf{Image source.} In this study, we selected the game \textit{Can You Escape?}~\cite{game} as our image source due to its close resemblance to real-world visuals. The image named ``scene1'' in each image set is the main scene, and the other images are details of this scene. In addition to visual realism, our benchmarks offer several significant advantages: 
1). Unified scene themes: While most existing datasets treat a single image as an independent scene, our benchmarks utilize multiple images with the same theme to create a cohesive and complete scene. 
2). Diverse task types: Our benchmarks encompass a variety of reasoning tasks within the same scenario, including reasoning about relationships between entities, interpreting digital codes, and rearranging sequences based on colors and shapes. 
3). Strong adaptability: We utilize advanced image processing tools to modify and augment images, a process that would be costly in real-world scenarios. This approach allows us to expand our benchmarks in a cost-effective manner.\\
\textbf{Question types.} In this paper, three question types are defined for \textbf{CVQA} and two for \textbf{CPVQA}. Specifically:
(1) In \textbf{CVQA}, we design three distinct types of questions to represent multi-graph reasoning tasks: props usage, password clues, prop collection. 
\begin{itemize}
    \item \textbf{Props usage}: In this task, the question explicitly identifies the tool in text form, prompting the model to use this prop to interact with a specific entity in the image. According to general real-world rules, in all scenarios, the tool can only interact with a unique entity within the given image.
    \item \textbf{Password clues}: In this task, the question provides a textual description that refers to an image, which acts as an index. The model is required to infer the corresponding image based on the given description, which may involve objects such as a digital password lock or entities arranged in random order.
    \item \textbf{Prop collection}:  In the this task, the questions act as guides, directing the model to identify two related entities within a scene composed of multiple images. In every image, this pair of related entities appears together, with one of them being an object that humans can physically interact with or pick up (e.g., a CD can be picked up, a stationary player cannot).
\end{itemize}
(2) In \textbf{CPVQA}, the image input task, specifically the password clues task in \textbf{CVQA}, primarily requires the model to identify relevant clues rather than producing precise outputs. However, by refining and standardizing benchmarks, we can further explore and assess the model’s capabilities in greater depth. In this study, we categorize the questions in the password clues task into two distinct types: password and sequence rearrangement.
\begin{itemize}
    \item \textbf{Password.} This task requires the model to infer a string of letters and numbers from a scene composed of multiple images. The question specifies the components (letters, numbers or a combination) and the number of digits in the final result. In some images, character combinations may appear in different forms, such as \texttt{Arabic numerals} or \texttt{Roman numerals}. It is important to note that in certain images, the final character sequence may need to be reorganized based on attributes like color, shape, or other distinguishing features.
    \item \textbf{Sequence rearrangement.} The task involves determining the correct order of entities in a given image (e.g., the order of the trophies in Figure~\ref{fig:problem_formulation}) within a scene composed of multiple images. The entities are encoded based on their order in the image from the question (e.g., `ABCD' from left to right). However, the final encoded sequence is influenced by information from other images, such as color, shape, or other attributes. To account for the possibility that the language model may repeat the question, we have designed the final true sequence to differ from the initial encoded sequence designed in the question.
\end{itemize}
\subsection{Benchmark Extension}
\label{subsec:extension method}
Existing image generation and automatic annotation methods are convenient but unreliable (e.g., GPT-generated text may be garbled). This error is critical for reasoning tasks, so we use manual annotation to expand benchmarks while ensuring data quality. This section  presents the scene expansion method, benchmark analysis, and score comparison, demonstrating its effectiveness.\\
\textbf{Scene extension.} In this study, we utilize \textbf{CVQA} and \textbf{CPVQA} as benchmarks to evaluate the reasoning ability of models in scenarios that involve multiple images and a single question. These benchmarks were designed with rigor in mind, emphasizing thematic consistency across scenarios and the provision of clear, unique answers. However, the benchmark’s limited size posed challenges to fully supporting our evaluation system. We utilize the image editing tool \texttt{Photoshop}~\cite{ps} to modify and expand the images.

\begin{figure}[ht]
    \centering
    \includegraphics[width=\linewidth]{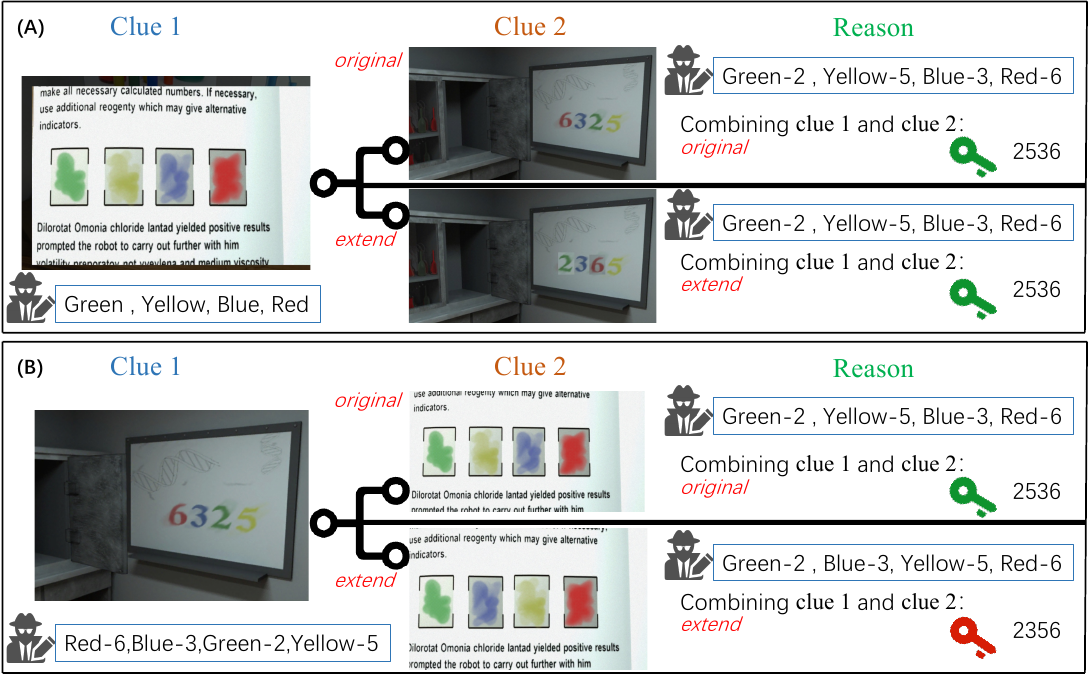}
    \caption{\textbf{Extension Methods.} Two examples of extension methods applied to the same scenario. In these examples, (A) indicates that \textit{expansion without altering the original answer}, while (B) indicates that \textit{expansion by altering the original answer}.}
    \label{fig:benchmark_example}
\end{figure}
\begin{figure}[ht]
    \centering
    \includegraphics[width=\linewidth]{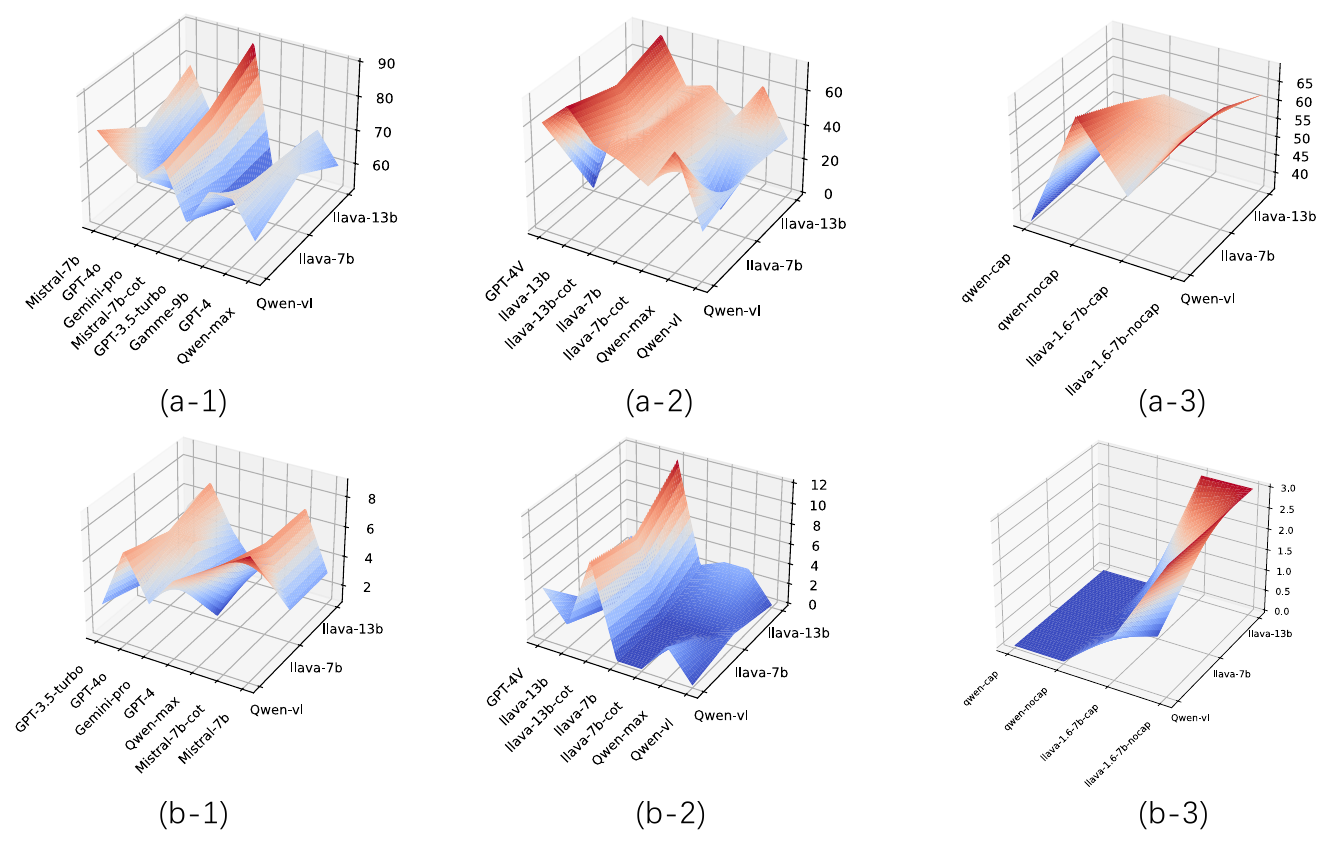}
    \caption{\textbf{Extension Method Validity.} Original benchmark \textbf{vs.} Extended benchmark: The number of differing results across various methods and models for the same task highlights the effectiveness of the extended benchmark, where `a' represents \textbf{CVQA}, and `b' represents \textbf{CPVQA}. Numbers `1', `2' and `3' correspond to the LLMs reasoning and MLLMs reasoning of method model inference~(Sec.~\ref{subsec:LLMs and MLLMs}) and semantic retrieval~(Sec.~\ref{subsec:semantic and visual retrieval}) respectively.}
    \label{fig:benchmark validity}
\end{figure}

Our expansion methods are categorized into two types: 
1) \textit{Expansion without altering the original answer}: In this approach, we rearrange the positions of the color blocks within the image without changing the original answer. Additionally, in some cases, we replace certain sequences that do not affect the final result, such as the symbols shown in Figure~\ref{fig:benchmark_example} (a).
2) \textit{Expansion by altering the original answer}: On one hand, sequences of characters and entities are modified by altering their order in certain samples. On the other hand, attributes such as color, shape, and other factors that indirectly influence the sequence of characters and entities in the image are adjusted, as illustrated in Figure~\ref{fig:benchmark_example} (b).\\
\textbf{Benchmark statistics.} In the \textbf{CVQA} benchmark, the base dataset consists of 522 images across 123 scenarios. By applying the scene expansion method, we doubled the dataset, resulting in 1,003 images and 227 scenarios. For the \textbf{CPVQA} benchmark, the base dataset contains 180 images and 32 scenarios. Using the same expansion method, we increased the dataset nearly fourfold, bringing the total to 661 images and 147 scenarios.\\
\textbf{Benchmark validity.} To verify the effectiveness of the extended benchmark in evaluating model capabilities, we compare the results of the extended benchmark with the original benchmark, utilizing various methods to answer questions across multiple models. 
The results are shown in Figure~\ref{fig:benchmark validity}, where the x-axis represents the captions from different sources, the $y$-axis represents the model, and the $z$-axis shows the number of differing answer scores for the same task in both the extended and original benchmarks (if both scores are 0, they are considered the same). For the same task across the two benchmarks, variations in score differences across models and methods indicate the effectiveness of our expansion approach in modifying scenarios.

%% file: sec/method.tex
\section{Methods}
\label{sec:method}
This section proposes three baseline methods to address our benchmarks without relying on prompt engineering. First, reasoning with models is discussed in Sec.\ref{subsec:LLMs and MLLMs}. Next, COT reasoning without prompts is covered in Sec.\ref{subsec:COT without prompting}. Finally, an approach for semantic retrieval on image sets is introduced in Sec.\ref{subsec:semantic and visual retrieval}. In this study, considering that most MLLMs~\cite{Mialon2023AugmentedLM} at this stage can effectively handle single-image input, all of methods are based on single-image inputs.
\subsection{Contextual Learning}
\label{subsec:LLMs and MLLMs}
In this subsection, we introduce two approaches for inference using LLMs and MLLMs.\\
\textbf{LLMs reasoning.} The benchmark we design consists of an image set with multiple images $V = \{v_1,v_2,\dots,v_n\}$ and a text-based question $T = (Q_{c} | Q_{p})$ as input. Since most current MLLMs effectively process only single images, captions are first generated for each image utilizing MLLMs. The scenario’s question, along with the captions from all images, is then input into LLMs for prediction:
\begin{equation}
    A_{LLM} = LLM\{T,\sum_{i=1}^n[Caption(v_i)]\}
\end{equation}
where $LLM\{T_1,T_2\}$ denotes that the LLM utilizes $T_1$ and $T_2$ as text inputs for prediction, generating the prediction result $A_{LLM}$, and $Caption(v)$ represents the conversion of image $v$ into caption utilizing MLLM.\\
\textbf{MLLMs reasoning.} In this study, each scene in the benchmark contains a main scene $V_m$ that, although not rich in detail, conveys the overall theme and key entities of the scene. Therefore, we first use the visual input from the $V_m$ to help MLLMs understand the theme of the scene, and then combine the detailed descriptions $\sum_{i=1}^n[Caption(v_i)]$ of individual images for the final prediction, specifically:
\begin{equation}
    A_{MLLM} = MLLM\{V_m,T,\sum_{i=1}^n[Caption(v_i)]\}
\end{equation}
where $MLLM\{v,T_m\}$ represents the MLLM utilizes $v$ as image input and $T_m$ as text input for prediction, generating the prediction result $A_{MLLM}$.
\subsection{COT Reasoning Without Prompting}
\label{subsec:COT without prompting}
This subsection primarily introduces a COT reasoning without prompting method. Specifically, we introduce the minimum margin decoding strategy to address uncertainty and then generate coherent answers.\\
\textbf{Minimum margin decoding.} Greedy decoding selects the best option at each step. Inspired by research~\cite{COTwithoutprompting}, the minimum margin decoding strategy not only considers the best option but also measures the model’s confidence by comparing the probability difference between the best and second-best options. Specifically, we calculate the probabilities of the best option $x1_t$ and the second-best option $x2_t$ at each step, utilizing their difference to represent the confidence: 
\begin{equation}
\Delta_t = p(x1_t | x_{<t}) - p(x2_t | x_{<t})
\end{equation}
where $t$ is each generation step and $p(x_t | x_{<t}$ represents the probability distribution of the model. The overall confidence of the final answer can be formalized as: 
\begin{equation}
\Delta_{\text{answer}} = \frac{1}{|x_{\text{answer}}|} \sum_{t \in \text{answer}} \Delta_t
\end{equation}
where $|x_{\text{answer}}|$ represents the total number of steps in the answer. A larger $\Delta_{\text{answer}}$ value indicates that the model has a higher confidence in the entire answer.\\
\textbf{Generate coherent answers.} In this study, when the model is uncertain at a given step $t$ and $\Delta_t$ falls below a certain threshold, we introduce randomness to ensure the consistency, credibility, and diversity of the generated results. Specifically, we use a temperature sampling method to introduce this randomness. The process is formulated as:
\begin{equation}
p(x_t | x_{<t}) = \frac{p(x_t | x_{<t})^{1/\mathcal{T}}}{\sum_{x' \in \mathcal{V}} p(x' | x_{<t})^{1/\mathcal{T}}}
\end{equation}
where $\mathcal{T}$ is a temperature parameter that controls the strength of randomness, and $\mathcal{V}$ is the vocabulary.
\subsection{Semantic Retrieval}
\label{subsec:semantic and visual retrieval}
In this subsection, we introduce a semantic retrieval method for handling multiple image inputs. While the main scene in contextual learning/~(Sec.~\ref{subsec:LLMs and MLLMs}) of this study provides the MLLM with key information, such as the theme of the scene as the image input, it may not always be the image most relevant to the answer. Therefore, across all known scenes, we evaluate the semantic relevance between the question and the scene descriptions to identify the image likely to be relevant to the answer. 
This study implements two methods:

(1)We utilize MLLM to generate captions $\mathcal{C}$ for complex scenes $V = \{v_1,v_2\dots,v_n\}$:
\begin{equation}
    \mathcal{C} = \sum_{i=1}^n[Caption(v_i)]
    \label{equ_captions}
\end{equation}
provide the LLM with $\mathcal{C}$ as background, and combine them with the scenario’s question. The LLM then determines the most semantically relevant image $v'$, which is considered the one most likely related to the answer. This image’s encoding is input into the MLLM as a visual embedding, along with the question and captions $\mathcal{C}$, for prediction.

(2) We utilize the language model’s semantic discernment capability to identify the caption $Caption(v')$ of the image $v'$ that is most likely relevant to the question, given $\mathcal{C}$ as background context in Equation~\ref{equ_captions}. After identifying this image $v'$, we input its encoding, together with the question, into the MLLM for prediction.

%% file: sec/experiments.tex
\section{Experiments}
\label{sec:experiments}
This section first introduces the details of exploratory experiment (Sec.~\ref{subsec:explori}), which utilizing the high-level reasoning model to perform combinatorial reasoning in complex scenarios. Next, we describe the experimental setup (Sec.~\ref{subsec:experiment setup}), which includes the benchmark evaluation mechanism and the models used in the experiments. Finally the results~(Sec.~\ref{subsec:main results}) and in-depth analysis (Sec.~\ref{subsec:experiment analysis}) of our proposed methods are presented.
\subsection{Exploratory Experiment}
\label{subsec:explori}
We test with manual prompts to stimulate more thoughtful model responses, as shown in Figure ~\ref{fig:intro_conformal}. The specific steps are as follows:
\begin{figure}[ht]
    \centering
    \includegraphics[width=\linewidth]{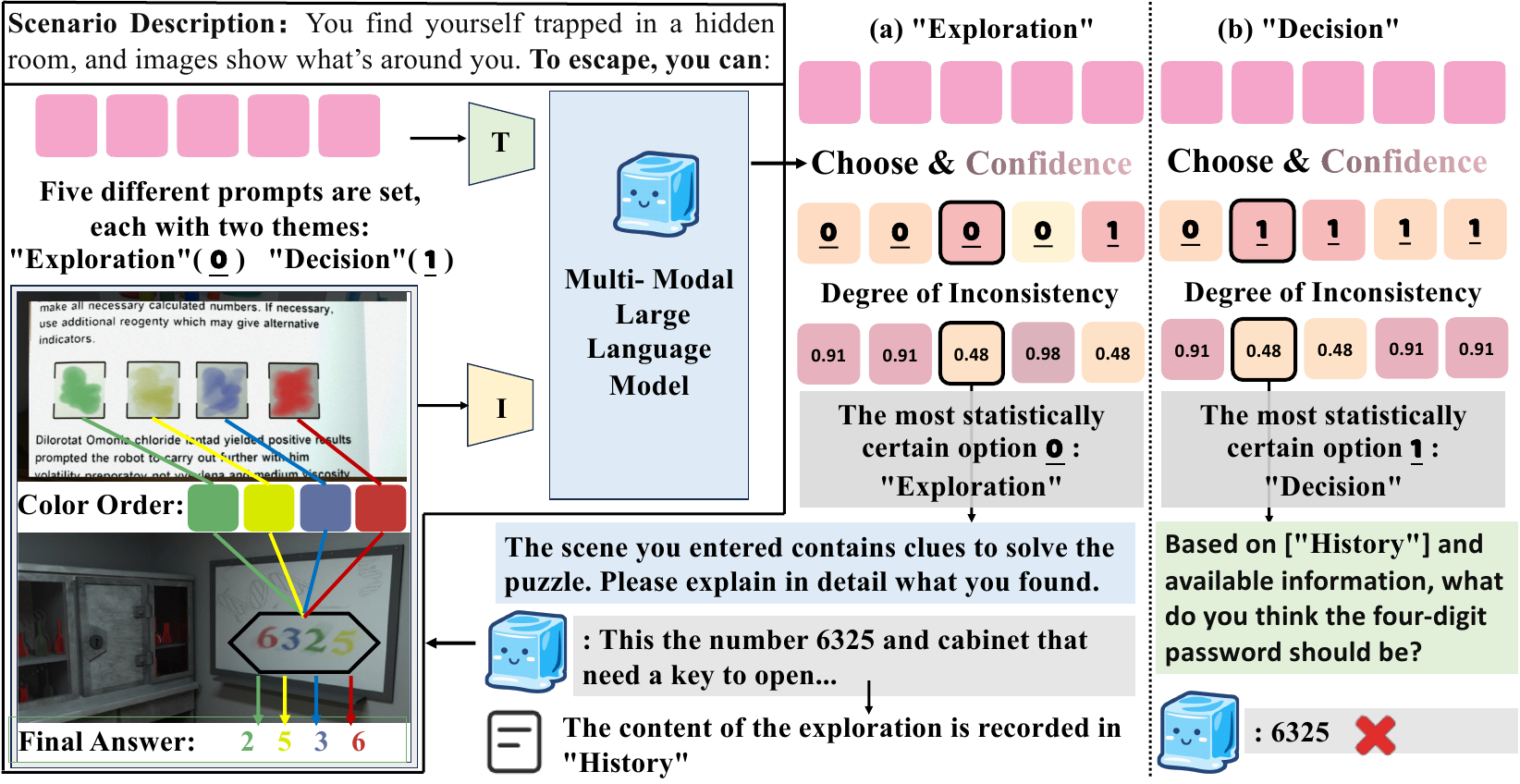}
    \caption{\textbf{Exploration Experiments.} Examples guide models to solve complex scenes through reasoning with manual prompts.}
    \label{fig:intro_conformal}
\end{figure}

(1) The model utilizes 5 different binary classification task prompts to make choices: `choose 0' indicates that the model should continue exploring a given scene $\mathcal{S}$, while `choose 1' signals that the model is ready to make a final action to attempt unveiling the mystery.

(2) A joint calibration set is constructed based on the original calibration set and the confidence interval from conformal prediction, allowing us to calculate degree of inconsistency: $p = \frac{ |\mathcal{D'}|+1}{|\mathcal{D}|+1}$, where $\mathcal{D'}$ represents samples in the calibration set $\mathcal{D}$ with confidence scores greater than or equal to the confidence of the current sample. A lower $p$ value indicates a higher confidence level.

(3) The choice with the lowest $p$ among the 5 prompts is utilized as the model’s most confident next action. If `choose 0' is selected, artificial prompts and ``previous actions'' are provided as text input, with $\mathcal{S}$ as the image input. $\mathcal{S}$ is then described in detail and summarized as context in ``previous actions''. If `choose 1' is selected, all scene images are utilized as image input, along with artificial prompts and ``previous actions'' as text input, to predict the final action needed to unveiling the mystery. As shown in Figure~\ref{fig:intro_conformal}, the model selects `choose 1' after first selecting `choose 0', enabling it to achieve step-by-step reasoning in the complex scenario over two steps.

We employ a high-level reasoning model ChatGPT-4o~\cite{chatgpt} to perform thoughtful reasoning across 36 complex scenarios. Surprisingly, none of the 36 responses are correct, suggesting that prompt engineering has minimal impact on this task, even with manually crafted prompts.
\subsection{Experimental Setup}
\label{subsec:experiment setup}
In this subsection, we first introduce the evaluation mechanism of our proposed benchmarks, followed by a detailed description of the models utilized in our experiments.\\
\textbf{Evaluation.} In this study, the evaluation mechanisms for \textbf{CVQA} and \textbf{CPVQA} differ. Specifically: (1) For \textbf{CVQA}, this paper require the final result to be a sentence that includes the image name and an interactive entity. We then calculate the number Ground Truth (GT) present in this sentence, denoted as $n_c$. If there are $x$ questions in total, the final score is formalized as:
\begin{equation}
    Score(CVQA) = \frac{\sum_1^x{\min(1,{n_c}/3})}{x}
\end{equation}
(2) For \textbf{CPVQA}, this study evaluates the final result as an ordered string defined by the question for each scenario (e.g., ‘ABCD’ from left to right, as shown in Figure~\ref{fig:problem_formulation}). The GT order is unique; if the predicted order is correct, the score is 1, otherwise, it is 0. If there are $x_1$ questions in total, the final score is formalized as:
\begin{equation}
    Score(CPVQA) \frac{\sum_{i=1}^{x_1} \delta_i}{x_1},\quad 
    \delta_i = \begin{cases}
    1, & \text{True} \\
    0, & \text{False}
    \end{cases}
\end{equation}\\
\textbf{Models.} 
In this paper, we utilize four open-source and six closed-source models to evaluate the performance of various models and configurations.
1). For caption generation, we utilize three open-source models: LLaVA-1.6-7b~\cite{LLaVA}, LLaVA-1.6-13b~\cite{LLaVA}, and Qwen-VL~\cite{Qwen-vl}. 2). For model reasoning tasks, we utilize open-source models such as Mistral-7b~\cite{Mistral7}, LLaVA-1.6-7b, LLaVA-1.6-13b, and Qwen-VL. Additionally, we utilize closed-source models including ChatGPT-3.5-turbo~\cite{chatgpt}, ChatGPT-4, ChatGPT-4o, ChatGPT-4V, Gemini Pro~\cite{Geminipro}, and Qwen-Max~\cite{Qwenmax}.
\subsection{Main Results}
\label{subsec:main results}
We first present the model score results for method contextual learning (Sec.~\ref{subsec:LLMs and MLLMs}) in Tables~\ref{tab:llm reasoning experiments} and~\ref{tab:mllm reasoning experiments}, and then present the score results of our improved methods—COT reasoning (Sec.~\ref{subsec:COT without prompting}) and semantic retrieval (Sec.~\ref{subsec:semantic and visual retrieval})—through ablation experiments shown in Tables~\ref{tab:cot without prompting} and~\ref{tab:semantic retrieval}, respectively.
\begin{table}[ht]
\resizebox{\linewidth}{!}{
\begin{tabular}{l|l|llll|lll}
\hline
\multicolumn{2}{c|}{\cellcolor[HTML]{ECF4FF}Benchmark}       & \multicolumn{4}{c|}{\cellcolor[HTML]{ECF4FF}CVQA(Score)/\%}                                            & \multicolumn{3}{c}{\cellcolor[HTML]{ECF4FF}CPVQA(Score)/\%}                    \\ \hline
Model         & Caption                & \cellcolor[HTML]{EFEFEF}all & \cellcolor[HTML]{EFEFEF}pc & \cellcolor[HTML]{EFEFEF}pu & \cellcolor[HTML]{EFEFEF}pcl & \cellcolor[HTML]{EFEFEF}all & \cellcolor[HTML]{EFEFEF}pas & \cellcolor[HTML]{EFEFEF}seq \\ \hline
ChatGPT-3.5-turbo & Qwen-VL       & {22.03} & {25.64} & {45.45} & {12.56} & {4.70} & {1.12} & {10.00} \\
ChatGPT-3.5-turbo & LLaVA-1.6-7b  & {20.70} & {23.08} & {44.24} & {11.41} & {2.68} & {1.12} & {5.00} \\
ChatGPT-3.5-turbo & LLaVA-1.6-13b & {21.73} & {37.18} & {46.06} & {9.82} & {3.36} & {1.12} & {6.67} \\ \hline
ChatGPT-4 &Qwen-VL &{29.52} &{\textbf{70.51}} &{51.53} &{13.93} &{6.04} &{1.12} &{\underline{13.33}}\\
ChatGPT-4 &LLaVA-1.6-7b &{30.84} &{\underline{67.95}} &{53.33} &{15.75} &{4.70} &{0} &{11.67}\\
ChatGPT-4 &LLaVA-1.6-13b &{28.05} &{\underline{67.95}} &{47.88} &{13.47} &{2.68} &{2.25} &{3.33}\\  \hline
ChatGPT-4o &Qwen-VL &{33.04} &{42.39} &{59.39} &{21.46} &{6.04} &{\underline{4.49}} &{8.33} \\
ChatGPT-4o &LLaVA-1.6-7b &{29.52} &{37.18} &{\textbf{60.61}} &{16.44} &{6.04} &{1.12} &{\underline{13.33}} \\
ChatGPT-4o &LLaVA-1.6-13b &{29.07} &{46.15} &{50.30} &{18.04} &{6.71} &{1.12} &{\textbf{15.00}} \\ \hline
Qwen-Max &Qwen-VL &{27.02} &{50.00} &{43.03} &{16.70} &{6.71} &{2.25} &{13.33}\\
Qwen-Max &LLaVA-1.6-7b &{24.67} &{35.90} &{50.91} &{12.79} &{5.37} &{\textbf{6.74}} &{3.33}\\
Qwen-Max &LLaVA-1.6-13b &{24.96} &{39.74} &{42.42} &{15.75} &{5.37} &{1.12} &{11.67}\\   \hline
Gemini Pro  &Qwen-VL &{26.28} &{25.64} &{28.48} &{25.57} &{4.70} &{1.12} &{10.00}\\
Gemini Pro  &LLaVA-1.6-7b &{27.75} &{29.49} &{31.52} &{26.03} &{\underline{7.38}} &{2.25} &{\textbf{15.00}}\\
Gemini Pro  &LLaVA-1.6-13b &{27.75} &{34.62} &{32.12} &{24.89} &{4.70} &{2.25} &{8.33}\\    \hline
Mistral-7b  &Qwen-VL &{\underline{36.12}} &{60.26} &{49.70} &{\underline{26.71}} &{\textbf{8.72}} &{\underline{4.49}} &{\textbf{15.00}}\\
Mistral-7b  &LLaVA-1.6-7b &{33.92} &{61.54} &{\underline{56.97}} &{20.32} &{4.70} &{1.12} &{10.00}\\
Mistral-7b  &LLaVA-1.6-13b &{\textbf{36.86}} &{64.10} &{47.88} &{\textbf{27.85}} &{6.04} &{1.12} &{\underline{13.33}}\\    \hline
\end{tabular}
}
\caption{\textbf{The main results of method LLMs reasoning} (Sec.~\ref{subsec:LLMs and MLLMs})\textbf{.} 
In the tables of this paper, the \textbf{bold font} is the first place in the same task category, and the \underline{underlined font} is the second place; `Caption' indicates the type of model that generates captions; and `all' represents the total score of all tasks in the benchmark, `pc' represents \textbf{p}rop \textbf{c}ollection , `pu' represents \textbf{p}rops \textbf{u}sage, `pcl' represents \textbf{p}assword \textbf{cl}ues, `pas' represents \textbf{pas}sword, and `seq' represents \textbf{seq}uence rearrangement.}
\label{tab:llm reasoning experiments}
\end{table}
\\\textbf{Results of LLMs reasoning. }Table~\ref{tab:llm reasoning experiments} presents the reasoning results of LLMs utilizing method LLM reasoning~(Sec.~\ref{subsec:LLMs and MLLMs}) on the \textbf{CVQA} and \textbf{CPVQA} benchmarks, showing that current models do not perform well on these tasks. First, the best results appear sequentially across most models, highlighting that no model excels completely at the tasks designed for our benchmarks. In \textbf{CVQA}, the `all' category represents the total score across the three task types, with the highest score reaching only 36.86\%, which shows no outstanding performance. In \textbf{CPVQA}, the highest score is 8.72\%. 
In summary, our findings indicate that the scoring results for utilizing LLMs in our proposed benchmarks are unsatisfactory, suggesting that LLMs alone struggle to perform combinatorial reasoning that integrates multiple perceptual inputs in complex scenes.
\begin{table}[ht]
\resizebox{\linewidth}{!}{
\begin{tabular}{l|l|llll|lll}
\hline
\multicolumn{2}{c|}{\cellcolor[HTML]{ECF4FF}Benchmark}       & \multicolumn{4}{c|}{\cellcolor[HTML]{ECF4FF}CVQA(Score)/\%}                                            & \multicolumn{3}{c}{\cellcolor[HTML]{ECF4FF}CPVQA(Score)/\%}                    \\ \hline
Model         & Caption                & \cellcolor[HTML]{EFEFEF}all & \cellcolor[HTML]{EFEFEF}pc & \cellcolor[HTML]{EFEFEF}pu & \cellcolor[HTML]{EFEFEF}pcl & \cellcolor[HTML]{EFEFEF}all & \cellcolor[HTML]{EFEFEF}pas & \cellcolor[HTML]{EFEFEF}seq \\ \hline
ChatGPT-4V  &Qwen-VL &{19.82} &{35.90} &{36.36} &{10.73} &{\underline{4.70}} &{\textbf{4.49}} &{\underline{5.00}}\\
ChatGPT-4V  &LLaVA-1.6-13b &{16.59} &{23.08} &{33.33} &{9.13} &{\textbf{5.37}} &{1.12} &{
\textbf{11.67}}\\    \hline
Qwen-Max    &Qwen-VL &{\textbf{27.90}} &{\underline{36.00}} &{\textbf{50.91}} &{17.81} &{2.68} &{1.12} &{\underline{5.00}}\\
Qwen-Max    &LLaVA-1.6-13b &{23.79} &{\textbf{38.46}} &{\underline{47.27}} &{12.33} &{0.67} &{0} &{1.67} \\ \hline
Qwen-VL     &Qwen-VL &{9.25} &{15.38} &{19.39} &{4.34} &{0.67} &{1.12} &{0}\\
Qwen-VL     &LLaVA-1.6-7b &{12,92} &{29.49} &{15.76} &{8.90} &{1.34} &{1.12} &{1.67}\\
Qwen-VL     &LLaVA-1.6-13b &{11.45} &{19.23} &{23.03} &{5.71} &{1.34} &{\underline{2.25}} &{0}\\ \hline
LLaVA-1.6-7b     &Qwen-VL &{16.30} &{19.23} &{30.91} &{10.27} &{0.67} &{0} &{1.67}\\
LLaVA-1.6-7b     &LLaVA-1.6-7b &{18.36} &{15.38} &{34.55} &{12.79} &{0} &{0} &{0}\\
LLaVA-1.6-7b     &LLaVA-1.6-13b &{21.29} &{25.64} &{32.73} &{16.21} &{1.67} &{\underline{2.25}} &{0}\\ \hline
LLaVA-1.6-13b     &Qwen-VL &{23.94} &{32.05} &{38.18} &{17.12} &{1.34} &{1.12} &{1.67}\\
LLaVA-1.6-13b     &LLaVA-1.6-7b &{23.94} &{30.77} &{31.52} &{\underline{19.18}} &{0.67} &{1.12} &{0}\\
LLaVA-1.6-13b     &LLaVA-1.6-13b &{\underline{26.14}} &{33.33} &{36.36} &{\textbf{21.00}} &{0.67} &{1.12} &{0}\\ \hline
\end{tabular}
}
\caption{\textbf{The main results of method MLLMs reasoning} (Sec.~\ref{subsec:LLMs and MLLMs}), and different from those of method LLMs reasoning in Table~\ref{tab:llm reasoning experiments} in that the main scene is utilized as the image input.}
\label{tab:mllm reasoning experiments}
\end{table}
\\\textbf{Results of MLLMs reasoning. }The main results of reasoning with the MLLM reasoning method~(Sec.~\ref{subsec:LLMs and MLLMs}) are shown in Table~\ref{tab:mllm reasoning experiments}. The closed-source model Qwen-max achieves the best performance in \textbf{CVQA} when captions are generated by Qwen-vl. However, when the main scene is provided to the model as the ``scene theme'', the overall performance of MLLM reasoning is lower than that of LLM reasoning. In \textbf{CVQA}, MLLM reasoning reaches only 27.90\% of the highest score, while LLM reasoning achieves up to 36.86\%. Similarly, in \textbf{CPVQA}, MLLM reasoning achieves only 5.37\% of the highest score, whereas LLM reasoning reaches 8.72\%. Additionally, when utilizing the same closed-source model, MLLM reasoning scores are generally lower than those of LLM reasoning; for instance, ChatGPT-4V performs lower than ChatGPT-4 in \textbf{CVQA}, and Qwen-max performs lower than Qwen-max without main scene input in \textbf{CPVQA}. This suggests that utilizing non-retrieved images as embeddings in complex scenes may weaken the model’s combinatorial reasoning ability. In summary, adding ``scene theme'' offers limited benefits for improving reasoning performance.
\begin{table}[ht]
\resizebox{\linewidth}{!}{
\begin{tabular}{l|l|c|llll|lll}
\hline
\multicolumn{3}{c|}{\cellcolor[HTML]{ECF4FF}Benchmark}       & \multicolumn{4}{c|}{\cellcolor[HTML]{ECF4FF}CVQA(Score)/\%}                                            & \multicolumn{3}{c}{\cellcolor[HTML]{ECF4FF}CPVQA(Score)/\%}                    \\ \hline
Model(LLMs)         & Caption    &COT             & \cellcolor[HTML]{EFEFEF}all & \cellcolor[HTML]{EFEFEF}pc & \cellcolor[HTML]{EFEFEF}pu & \cellcolor[HTML]{EFEFEF}pcl & \cellcolor[HTML]{EFEFEF}all & \cellcolor[HTML]{EFEFEF}pas & \cellcolor[HTML]{EFEFEF}seq \\ \hline
Mistral-7b &Qwen-VL & &{36.12} &{60.26} &{49.70} &{26.71} &{8.72} &\cellcolor[HTML]{fea3a0}{4.49} &{15.00} \\
Mistral-7b &Qwen-VL &\checkmark &\cellcolor[HTML]{fea3a0}{55.21} &\cellcolor[HTML]{fea3a0}{70.50} &\cellcolor[HTML]{fea3a0}{64.85} &\cellcolor[HTML]{fea3a0}{48.86} &\cellcolor[HTML]{fea3a0}{10.07} &{3.37} &\cellcolor[HTML]{fea3a0}{20.00} \\ \hdashline 
Mistral-7b &LLaVA-1.6-7b & &{33.92} &{61.54} &{56.97} &{20.32} &{4.70} &{1.12} &{10.00} \\
Mistral-7b &LLaVA-1.6-7b &\checkmark &\cellcolor[HTML]{fea3a0}{49.19} &\cellcolor[HTML]{fea3a0}{69.23} &\cellcolor[HTML]{fea3a0}{72.12} &\cellcolor[HTML]{fea3a0}{36.99} &\cellcolor[HTML]{fea3a0}{9.40} &\cellcolor[HTML]{fea3a0}{6.74} &\cellcolor[HTML]{fea3a0}{13.33} \\ \hdashline 
Mistral-7b &LLaVA-1.6-13b & &{36.86} &{64.10} &{47.88} &{27.85} &\cellcolor[HTML]{fea3a0}{6.04} &{1.12} &\cellcolor[HTML]{fea3a0}{13.33} \\
Mistral-7b &LLaVA-1.6-13b &\checkmark &\cellcolor[HTML]{fea3a0}{51.40} &\cellcolor[HTML]{fea3a0}{73.08} &\cellcolor[HTML]{fea3a0}{64.24} &\cellcolor[HTML]{fea3a0}{42.69} &{4.03} &\cellcolor[HTML]{fea3a0}{2.25} &{6.67} \\ \hline 
Model(MLLMs)         & Caption    &COT             & \cellcolor[HTML]{EFEFEF}all & \cellcolor[HTML]{EFEFEF}col & \cellcolor[HTML]{EFEFEF}use & \cellcolor[HTML]{EFEFEF}pas & \cellcolor[HTML]{EFEFEF}all & \cellcolor[HTML]{EFEFEF}num & \cellcolor[HTML]{EFEFEF}ord \\ \hline
LLaVA-1.6-7b &Qwen-VL & &\cellcolor[HTML]{fea3a0}{16.30} &{19.23} &\cellcolor[HTML]{fea3a0}{30.91} &\cellcolor[HTML]{fea3a0}{10.27} &\cellcolor[HTML]{fea3a0}{0.67} &{0} &\cellcolor[HTML]{fea3a0}{1.67}\\
LLaVA-1.6-7b &Qwen-VL &\checkmark &{11.75} &\cellcolor[HTML]{fea3a0}{23.08} &{22.42} &{5.71} &{0} &{0} &{0}\\ \hdashline
LLaVA-1.6-7b &LLaVA-1.6-7b & &\cellcolor[HTML]{fea3a0}{18.36} &\cellcolor[HTML]{fea3a0}{15.38} &\cellcolor[HTML]{fea3a0}{34.55} &\cellcolor[HTML]{fea3a0}{12.79} &{0} &{0} &{0}\\
LLaVA-1.6-7b &LLaVA-1.6-7b &\checkmark &{15.71} &{14.10} &{31.52} &{10.46} &{0} &{0} &{0}\\ \hdashline
LLaVA-1.6-7b &LLaVA-1.6-13b & &\cellcolor[HTML]{fea3a0}{21.29} &\cellcolor[HTML]{fea3a0}{25.64} &\cellcolor[HTML]{fea3a0}{32.73} &\cellcolor[HTML]{fea3a0}{16.21} &\cellcolor[HTML]{fea3a0}{1.67} &\cellcolor[HTML]{fea3a0}{2.25} &{0}\\
LLaVA-1.6-7b &LLaVA-1.6-13b &\checkmark &{13.95} &{16.67} &{29.70} &{7.53} &{0.67} &{1.12} &{0}\\ \hline
LLaVA-1.6-13b &Qwen-VL & &\cellcolor[HTML]{fea3a0}{23.94} &\cellcolor[HTML]{fea3a0}{32.05} &\cellcolor[HTML]{fea3a0}{38.18} &\cellcolor[HTML]{fea3a0}{17.12} &{1.34} &{1.12} &{1.67}\\
LLaVA-1.6-13b &Qwen-VL &\checkmark &{16.59} &{21.79} &{27.88} &{11.42} &\cellcolor[HTML]{fea3a0}{14.09} &\cellcolor[HTML]{fea3a0}{3.37} &\cellcolor[HTML]{fea3a0}{30.00}\\ \hdashline
LLaVA-1.6-13b &LLaVA-1.6-7b & &\cellcolor[HTML]{fea3a0}{23.94} &\cellcolor[HTML]{fea3a0}{30.77} &\cellcolor[HTML]{fea3a0}{31.52} &\cellcolor[HTML]{fea3a0}{19.18} &{0.67} &{1.12} &{0}\\
LLaVA-1.6-13b &LLaVA-1.6-7b &\checkmark &{14.39} &{26.92} &{19.39} &{10.27} &\cellcolor[HTML]{fea3a0}{16.78} &\cellcolor[HTML]{fea3a0}{8.98} &\cellcolor[HTML]{fea3a0}{28.33}\\ \hdashline
LLaVA-1.6-13b &LLaVA-1.6-13b & &\cellcolor[HTML]{fea3a0}{26.14} &\cellcolor[HTML]{fea3a0}{33.33} &\cellcolor[HTML]{fea3a0}{36.36} &\cellcolor[HTML]{fea3a0}{21.00} &{0.67} &{1.12} &{0}\\
LLaVA-1.6-13b &LLaVA-1.6-13b &\checkmark &{13.51} &{29.49} &{20.61} &{7.99} &\cellcolor[HTML]{fea3a0}{16.11} &\cellcolor[HTML]{fea3a0}{10.11} &\cellcolor[HTML]{fea3a0}{25.00}\\ \hline
\end{tabular}
}
\caption{\textbf{The ablation results of method COT reasoning without prompting} (Sec.~\ref{subsec:COT without prompting}). \textcolor[HTML]{fea3a0}{Red areas} indicate better results with or without COT reasoning, `\checkmark' denotes taking COT reasoning.}
\label{tab:cot without prompting}
\end{table}
\\\textbf{Ablation on COT reasoning without prompting.} In Table~\ref{tab:cot without prompting}, we utilize the ablation method to demonstrate the effect of COT reasoning without prompting in method COT reasoning without prompting~(Sec.~\ref{subsec:COT without prompting}). 1) In the LLM reasoning method, the open-source model Mistral-7b was utilized as the base model. Table~\ref{tab:cot without prompting} presents the COT reasoning method significantly improved the model’s performance on the \textbf{CVQA} benchmark. However, it only partially enhance performance on the \textbf{CPVQA} benchmark. 2) In the MLLM reasoning method, this paper explores models with different parameter sizes of LLaVA-1.6. We find that in \textbf{CVQA}, the COT reasoning method had little to no effect on improving the performance of MLLMs. In \textbf{CPVQA}, COT reasoning method led to greater improvements for models with larger parameters (13b), which may be related to the random design of our COT reasoning method. Since the number of parameters is proportional to the vocabulary size, models with more parameters have a higher likelihood of generating a complex COT reasoning path. The COT reasoning method may be more effective for text reasoning and models with larger parameters, due to its inherent randomness.
\begin{table}[ht]
\resizebox{\linewidth}{!}{
\begin{tabular}{l|l|c|llll|lll}
\hline
\multicolumn{3}{c|}{\cellcolor[HTML]{ECF4FF}Benchmark}       & \multicolumn{4}{c|}{\cellcolor[HTML]{ECF4FF}CVQA(Score)/\%}                                            & \multicolumn{3}{c}{\cellcolor[HTML]{ECF4FF}CPVQA(Score)/\%}                    \\ \hline
Model         & Caption    &DOI             & \cellcolor[HTML]{EFEFEF}all & \cellcolor[HTML]{EFEFEF}pc & \cellcolor[HTML]{EFEFEF}pu & \cellcolor[HTML]{EFEFEF}pcl & \cellcolor[HTML]{EFEFEF}all & \cellcolor[HTML]{EFEFEF}pas & \cellcolor[HTML]{EFEFEF}seq\\ \hline
LLaVA-1.6-7b &Qwen-VL &- &{16.30} &{19.23} &{30.91} &{10.27} &{0.67} &{0} &{1.67} \\ \hdashline
LLaVA-1.6-7b &Qwen-VL &\checkmark &{18.21} &\cellcolor[HTML]{fea3a0}{25.64} &{33.94} &{10.96} &\cellcolor[HTML]{fea3a0}{2.01} &\cellcolor[HTML]{fea3a0}{2.25} &{1.67} \\ \hdashline
LLaVA-1.6-7b &Qwen-VL &\ding{55} &\cellcolor[HTML]{fea3a0}{26.58} &{21.79} &\cellcolor[HTML]{fea3a0}{49.07} &\cellcolor[HTML]{fea3a0}{18.72} &\cellcolor[HTML]{fea3a0}{2.01} &{0} &\cellcolor[HTML]{fea3a0}{5.00} \\ \hline
LLaVA-1.6-7b &LLaVA-1.6-7b &- &{18.36} &{15.38} &{34.55} &{12.79} &{0} &{0} &{0} \\ \hdashline
LLaVA-1.6-7b &LLaVA-1.6-7b &\checkmark &{19.24} &\cellcolor[HTML]{fea3a0}{23.08} &{35.15} &{12.56} &{0.67} &{0} &{1.67} \\ \hdashline
LLaVA-1.6-7b &LLaVA-1.6-7b &\ding{55} &\cellcolor[HTML]{fea3a0}{26.28} &{21.79} &\cellcolor[HTML]{fea3a0}{52.73} &\cellcolor[HTML]{fea3a0}{17.12} &\cellcolor[HTML]{fea3a0}{2.68} &\cellcolor[HTML]{fea3a0}{1.12} &\cellcolor[HTML]{fea3a0}{5.00} \\ \hline
LLaVA-1.6-7b &LLaVA-1.6-13b &- &\cellcolor[HTML]{fea3a0}{23.94} &\cellcolor[HTML]{fea3a0}{32.05} &{38.18} &\cellcolor[HTML]{fea3a0}{17.12} &{1.34} &\cellcolor[HTML]{fea3a0}{1.12} &{1.67} \\ \hdashline
LLaVA-1.6-7b &LLaVA-1.6-13b &\checkmark &{19.82} &\cellcolor[HTML]{fea3a0}{32.05} &{33.94} &{12.33} &{0.67} &\cellcolor[HTML]{fea3a0}{1.12} &{0} \\ \hdashline
LLaVA-1.6-7b &LLaVA-1.6-13b &\ding{55} &{23.35} &{24.36} &\cellcolor[HTML]{fea3a0}{50.30} &{13.01} &\cellcolor[HTML]{fea3a0}{2.01} &{0} &\cellcolor[HTML]{fea3a0}{5.00} \\ \hline
Qwen-VL &Qwen-VL &- &{9.25} &{15.38} &{19.39} &{4.34} &{0.67} &\cellcolor[HTML]{fea3a0}{1.12} &{0} \\ \hdashline
Qwen-VL &Qwen-VL &\checkmark &{10.13} &{11.54} &{23.03} &{5.02} &{0} &{0} &{0} \\ \hdashline
Qwen-VL &Qwen-VL &\ding{55} &\cellcolor[HTML]{fea3a0}{20.70} &\cellcolor[HTML]{fea3a0}{20.51} &\cellcolor[HTML]{fea3a0}{41.82} &\cellcolor[HTML]{fea3a0}{12.79} &\cellcolor[HTML]{fea3a0}{1.34} &\cellcolor[HTML]{fea3a0}{1.12} &\cellcolor[HTML]{fea3a0}{1.67} \\ \hline
Qwen-VL &LLaVA-1.6-7b &- &{12,92} &\cellcolor[HTML]{fea3a0}{29.49} &{15.76} &{8.90} &\cellcolor[HTML]{fea3a0}{1.34} &\cellcolor[HTML]{fea3a0}{1.12} &\cellcolor[HTML]{fea3a0}{1.67} \\ \hdashline
Qwen-VL &LLaVA-1.6-7b &\checkmark &{12.92} &{20.51} &{23.03} &{7.76} &{0.67} &{0} &\cellcolor[HTML]{fea3a0}{1.67} \\ \hdashline
Qwen-VL &LLaVA-1.6-7b &\ding{55} &\cellcolor[HTML]{fea3a0}{22.76} &{25.64} &\cellcolor[HTML]{fea3a0}{45.45} &\cellcolor[HTML]{fea3a0}{13.70} &\cellcolor[HTML]{fea3a0}{1.34} &\cellcolor[HTML]{fea3a0}{1.12} &\cellcolor[HTML]{fea3a0}{1.67} \\ \hline
Qwen-VL &LLaVA-1.6-13b &- &{11.45} &\cellcolor[HTML]{fea3a0}{19.23} &{23.03} &{5.71} &{1.34} &{2.25} &{0} \\ \hdashline
Qwen-VL &LLaVA-1.6-13b &\checkmark &{12.92} &\cellcolor[HTML]{fea3a0}{26.92} &{24.24} &{6.16} &\cellcolor[HTML]{fea3a0}{2.01} &\cellcolor[HTML]{fea3a0}{3.37} &{0} \\ \hdashline
Qwen-VL &LLaVA-1.6-13b &\ding{55} &{19.53} &{24.36} &\cellcolor[HTML]{fea3a0}{42.42} &\cellcolor[HTML]{fea3a0}{10.05} &{1.34} &{1.12} &\cellcolor[HTML]{fea3a0}{1.67} \\ \hline
\end{tabular}
}
\caption{\textbf{The ablation results of method semantic retrieval} (Sec.~\ref{subsec:semantic and visual retrieval}), where the DOI indicates Description of Other Images. The `DOI' in the table, (\checkmark) combines other image captions in the final prediction; (\ding{55}) excludes combination, aligning with semantic retrieval (Sec.~\ref{subsec:semantic and visual retrieval}); (-) uses only MLLM for reasoning.}
\label{tab:semantic retrieval}
\end{table}
\\\textbf{Ablation on semantic retrieval.} We compare the results of three methods in Table~\ref{tab:semantic retrieval}, the MLLM reasoning method in contextual learning (Sec.~\ref{subsec:LLMs and MLLMs}), utilizing inputs $\mathcal{C}$ and $v'$, and only $v'$ as input, without $\mathcal{C}$ (Sec.~\ref{subsec:semantic and visual retrieval}). Specifically, in \textbf{CVQA}, the method that relies solely on $v'$ as input, without incorporating $\mathcal{C}$, demonstrates a significant advantage. Notably, when utilizing captions generated by Qwen-VL and employing Qwen-VL for reasoning, this approach consistently achieves the highest scores. In contrast, the method utilizing both $\mathcal{C}$ and $v'$ yields suboptimal results, suggesting that in \textbf{CVQA}, our approach in semantic retrieval (Sec.~\ref{subsec:semantic and visual retrieval}) outperforms contextual learning (Sec.~\ref{subsec:LLMs and MLLMs}). Similarly, in \textbf{CPVQA}, the method that relies solely on $v'$ without incorporating $\mathcal{C}$ demonstrates a slight performance advantage over other approaches, whereas the MLLM reasoning method within the contextual learning framework produces less than optimal results.\\
\textbf{Additional Experiments. }
\label{AppendixB}
Given that models capable of handling multi-image inputs exist. To maintain the comprehensiveness of our experiment, we utilize multi-image input models to directly evaluate their performance on our proposed benchmarks. The results are presented in Table~\ref{tab:mut-model}. 
\begin{table}[ht]
\resizebox{\linewidth}{!}{
\begin{tabular}{ll|llll|lll}
\hline
\multicolumn{2}{c|}{Model}      & all(\textbf{CVQA})   & pc    & pu    & pcl   & all(\textbf{CPVQA})   & pas   & seq   \\ \hline
\multicolumn{2}{l|}{LLaVA-OV-7b~\cite{Li2024LLaVAOneVisionEV}}            & 24.08 & \underline{35.89} & \textbf{40.00} & 15.98 & 0.45   & \underline{6.67}  & 0  \\
\multicolumn{2}{l|}{Qwen2-vl-7B~\cite{Qwen2VL}}                & 18.94 & 20.51 & 24.24 & 16.67 & 2.91    & 3.88   & 2.25   \\
\multicolumn{2}{l|}{Qwen-vl-max~\cite{Qwenmax}}                & \underline{30.10} & 30.77 & \underline{26.06} & \underline{31.51} & \underline{3.58}    & 3.37   & \underline{3.89}   \\
\multicolumn{2}{l|}{Qvq-72B-preview~\cite{qvq-72b-preview}}            & \textbf{46.70} & \textbf{43.59} & \textbf{40.00} & \textbf{49.77} & \textbf{20.36}   & \textbf{26.58}  & \textbf{11.11}  \\
 \hline
\end{tabular}
}
\caption{\textbf{Evaluation on multiple image inputs.} Performance of models with multiple images as input on benchmarks.}
\label{tab:mut-model}
\end{table}
Even the large-parameter model Qwen-vl-max achieves an overall performance of only 30.10\% on our \textbf{CVQA} benchmark, and more shockingly, it scores just 3.58\% on the \textbf{CPVQA}. However, while the score of Qvq-72B is not particularly high, it outperforms other models and methods.\\
\subsection{In-depth Analysis}
\label{subsec:experiment analysis}
In our exploration and experiments, we identify three primary types of errors in the model’s reasoning. First, the model often relies on a single source of information rather than integrating multiple relevant inputs for joint reasoning. For instance, it may generate the answer ``6325'' without considering the color order depicted in Figure~\ref{fig:intro_example}, highlighting deficiencies in its multimodal retrieval and integration capabilities. Second, the model tends to over-rely on the explicit content of the query without critically analyzing additional contextual information. For example, in Figure ~\ref{fig:problem_formulation}, it directly concludes that the final trophy sequence is ``ABCD'' reflecting a tendency to follow surface-level cues rather than engage in deeper reasoning. Finally, when faced with ambiguous or incomplete information, the model often determines that the provided data is insufficient and instead resorts to prior knowledge for inference. This over-reliance on pre-existing biases underscores its limitations in fine-grained recognition and precise analytical reasoning.\\
Furthermore, effective methods, as shown in Tables~\ref{tab:cot without prompting} and~\ref{tab:mut-model}, models using COT or trained on COT data perform better, suggesting that enhanced reasoning methods, such as COT~\cite{COTwithoutprompting,Wei2022ChainOT} and RL-trained models~\cite{DeepSeekAI2025DeepSeekR1IR}, may be more effective in analyzing information from multiple sources. Another notable observation is that text-based reasoning frequently outperforms joint reasoning with the input image, indicating that the integration of multiple visual information sources warrants further refinement.

%% file: sec/conclusion.tex
\section{Conclusion}
\label{sec:conclusion}
We present two benchmarks to assess models’ combinatorial reasoning with multiple cues in complex scenarios. Experiments reveal current models struggle. We propose three baselines that mitigates these limitations, and guides future research. This study reveal key areas for improvement in multimodal reasoning, contextual comprehension, and fine-grained perception, which are crucial for enhancing the model’s robustness in complex scenarios.\\
\textit{Limitations:} Due to the dataset’s complexity, we rely on manual annotations to ensure methodological rigor. We anticipate that future research will improve scalability.

%% file: sec/appendix.tex
\maketitlesupplementary
\appendix
\section{Benchmark Solution Formats}
\label{AppendixA}
In this Section, in addition to the solved forms of \textbf{CVQA} and \textbf{CPVQA} shown in Figure~\ref{fig:problem_formulation}, we additionally show three other tasks of \textbf{CVQA} and \textbf{CPVQA} to emphasize the characteristics of different tasks on different benchmarks. This paper specifically quotes these images and explains them in text in Sec.~\ref{subsec:CVQA} and Sec.~\ref{subsec:CPVQA}. 
\begin{figure}[h]
    \centering
    \includegraphics[width=\linewidth]{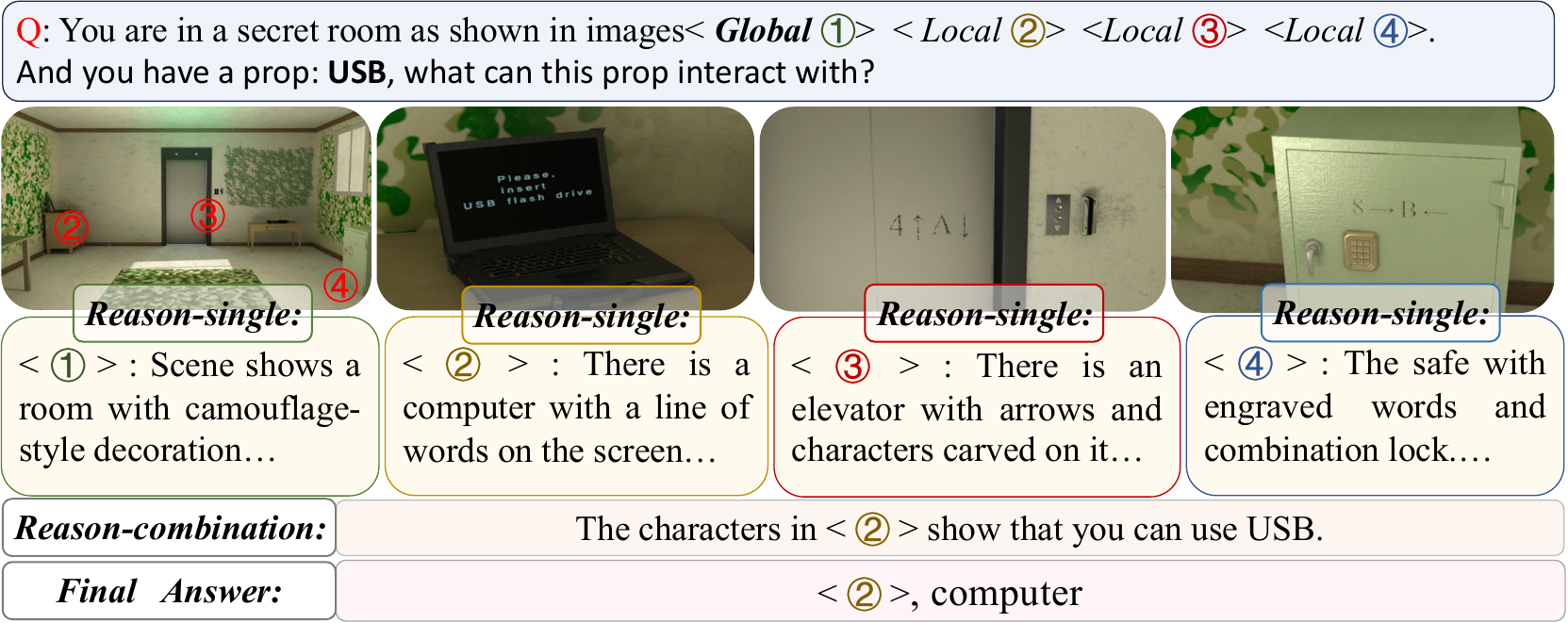}
    \caption{\textbf{Solution Formats.} The solution format for the prop usage task in \textbf{CVQA}.}
    \label{fig:figappendix1}
\end{figure}
\begin{figure}[h]
    \centering
    \includegraphics[width=\linewidth]{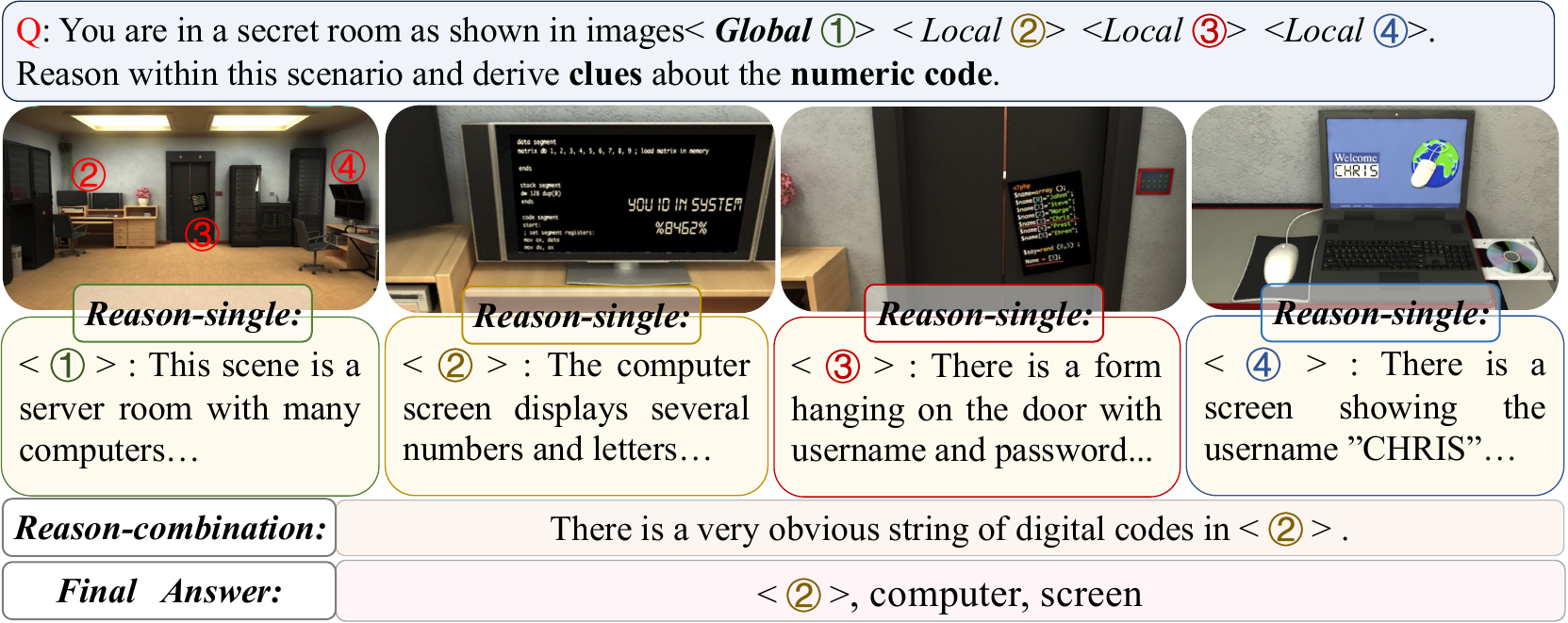}
    \caption{\textbf{Solution Formats.} The format for solutions to the password clues task in \textbf{CVQA}.}
    \label{fig:figappendix2}
\end{figure}
\begin{figure}[h]
    \centering
    \includegraphics[width=\linewidth]{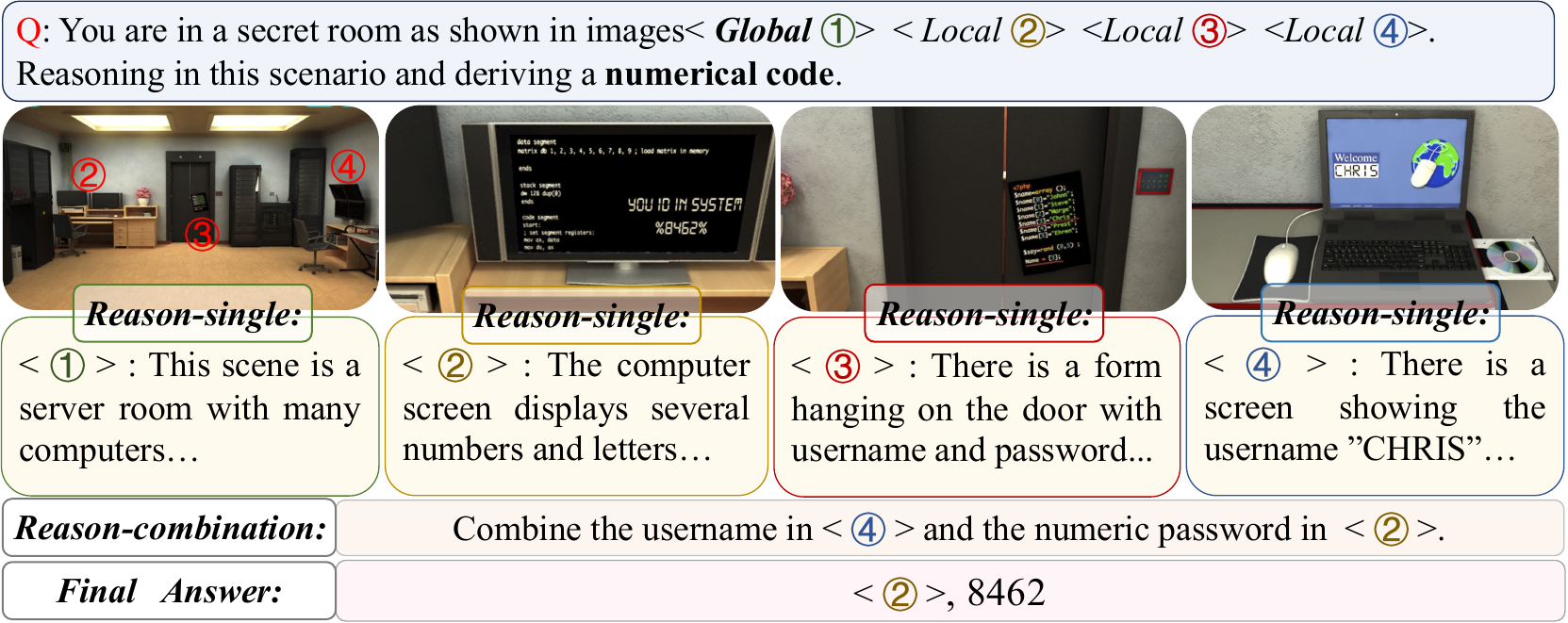}
    \caption{\textbf{Solution Formats.} The format used for providing solutions to the passwords task in \textbf{CPVQA}.}
    \label{fig:figappendix3}
\end{figure}
\input{sec/related_work}

%% file: sec/related_work.tex
\section{Related Work}
\label{sec:related work}
\textbf{Reasoning with LLMs.} LLMs~\cite{Vicuna,llama,internLM} are increasingly used in various reasoning tasks, including mixed reasoning~\cite{Yu2020ReClorAR}, arithmetic reasoning~\cite{gsm8k,GSMSymbolic}, and deductive reasoning~\cite{ProOnto}. Earlier research~\cite{ftforreasoning,ftforreasoning2,Yang2022GeneratingNL} has enhanced the reasoning abilities of LLMs through fine-tuning, but later studies~\cite{fewshotlearning,Emergent} revealed that fine-tuning can be both costly and less effective compared to using contextual examples. More recently, several strategies have been developed to leverage in-context Learning~\cite{Dong2022ASO,Bertsch2024InContextLW} and prompt design~\cite{Amatriain2024PromptDA,Shinn2023ReflexionLA,relatedprompt} methods, such as COT prompt~\cite{Wei2022ChainOT,Kojima2022LargeLM}, incremental reasoning~\cite{Zheng2023ProgressiveHintPI}, and Tree-of-Thought prompt~\cite{Yao2023TreeOT}. Previous methods have made significant contributions to enhancing model reasoning abilities; however, they primarily focus on individual reasoning within scenarios. In contrast, this paper aims to explore and address the combinatorial reasoning abilities of LLMs in complex scenarios.

\noindent\textbf{Benchmark.} Previous traditional vision-language benchmarks have primarily focused on specific tasks, such as visual commonsense reasoning~\cite{visualcommon,Wang2020VisualCR}, textual reasoning based on visual information~\cite{textvqa,He2021VisualSA}, logical question reasoning~\cite{CLEVR}, and other vision-language tasks~\cite{nextqa,videoqa}. 
Most of these benchmarks~\cite{okvqa,aokvqa,GVQA,vqav2} are designed in simple scenarios, lacking consideration for complex scenarios. This paper contends that, while existing benchmarks~\cite{mme,seedbench,mmbench,multiimages1,multiimages2,Plummer2015Flickr30kEC} are valuable for synthesizing vision-language capabilities, there is a need to focus on tasks in complex scenarios to promote continuous advancements in AI systems. Specifically, our benchmark offers advantages such as a detailed classification of question types and a focus on constructing benchmarks in complex scenes.
